\documentclass[10pt,twocolumn,letterpaper]{article}
\usepackage{tikz}
\usepackage{pgfplots}
\usepackage{verbatim}
\usepackage{xcolor}
\usepackage{cvpr}
\usepackage{times}
\usepackage{epsfig}
\usepackage{graphicx}
\usepackage{amsmath}
\usepackage{amssymb}
\usepackage{float}
\usepackage[linesnumbered,ruled]{algorithm2e}

\cvprfinalcopy 
\def\cvprPaperID{1542}

\begin{document}

\title{Rectifying Self Organizing Maps \\ for Automatic Concept Learning from Web Images}

\author{Eren Golge\\
Bilkent University\\
06800 Ankara/Turkey\\
{\tt\small eren.golge@bilkent.edu.tr}
\and
Pinar Duygulu\\
Bilkent University\\
06800 Ankara/Turkey\\
{\tt\small pinar.duygulu@gmail.com}
}
\maketitle

\begin{abstract}
We attack the problem of learning concepts automatically from noisy web image search results. Going beyond low level attributes, such as colour and texture, we explore weakly-labelled datasets for the learning of higher level concepts, such as scene categories. 
The idea is based on discovering common characteristics shared among subsets of images by posing a method that is able to organise the data while eliminating irrelevant instances.  We propose a novel clustering and outlier detection method, namely Rectifying Self Organizing Maps (RSOM). 
Given an image collection returned for a concept query, RSOM provides clusters pruned from outliers. Each cluster is used to train a model representing a different characteristics of the concept. The proposed method outperforms the state-of-the-art studies on the task of learning low-level concepts,  and it is competitive in learning higher level concepts as well. It is capable to work at large scale with no supervision through exploiting the available sources. 
\end{abstract}
\vspace{-0.5cm}

\section{Introduction}
The need for manually labelled data continues to be one of the most important limitations in large scale object/scene recognition.
Recently, use of visual attributes have become attractive as being helpful in describing properties shared by multiple categories and resulting in novel  category recognition. owever, most of the methods require learning of visual attributes from labelled data, and cannot eliminate human effort. Yet, it may be more difficult to describe an attribute than an object, and localisation may not be trivial. 

\begin{figure}
\begin{center}
\begin{tabular}{ccccc}
\includegraphics[width=0.14\linewidth, height=1cm]{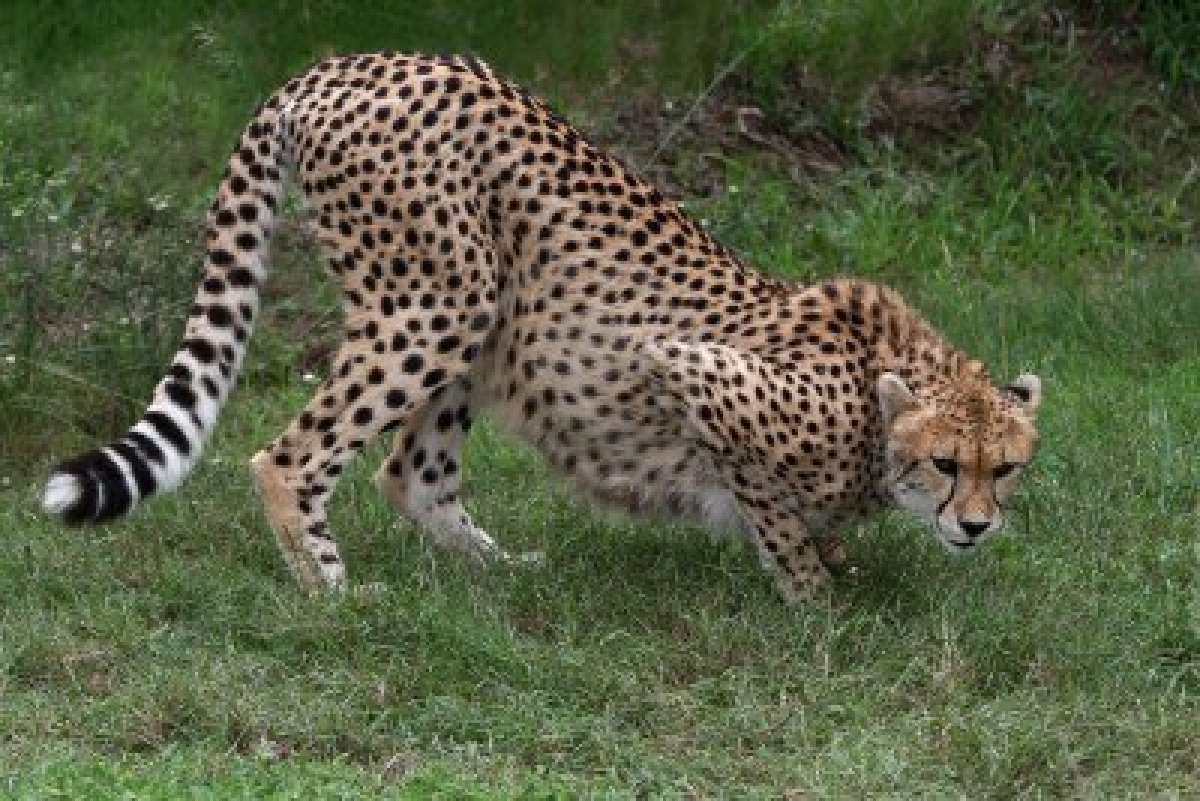} &
\includegraphics[width=0.14\linewidth, height=1cm]{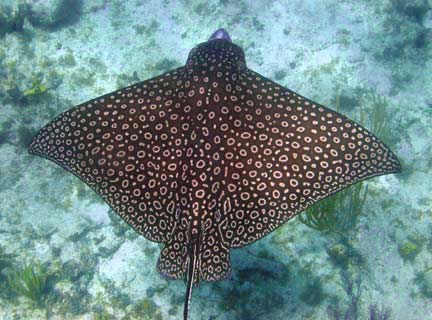} &
\includegraphics[width=0.14\linewidth, height=1cm]{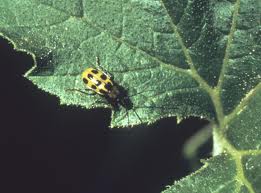} &
\includegraphics[width=0.14\linewidth, height=1cm]{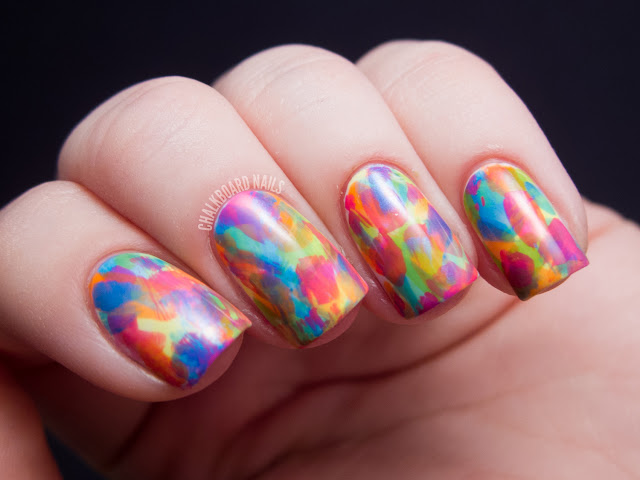} &
\includegraphics[width=0.14\linewidth, height=1cm]{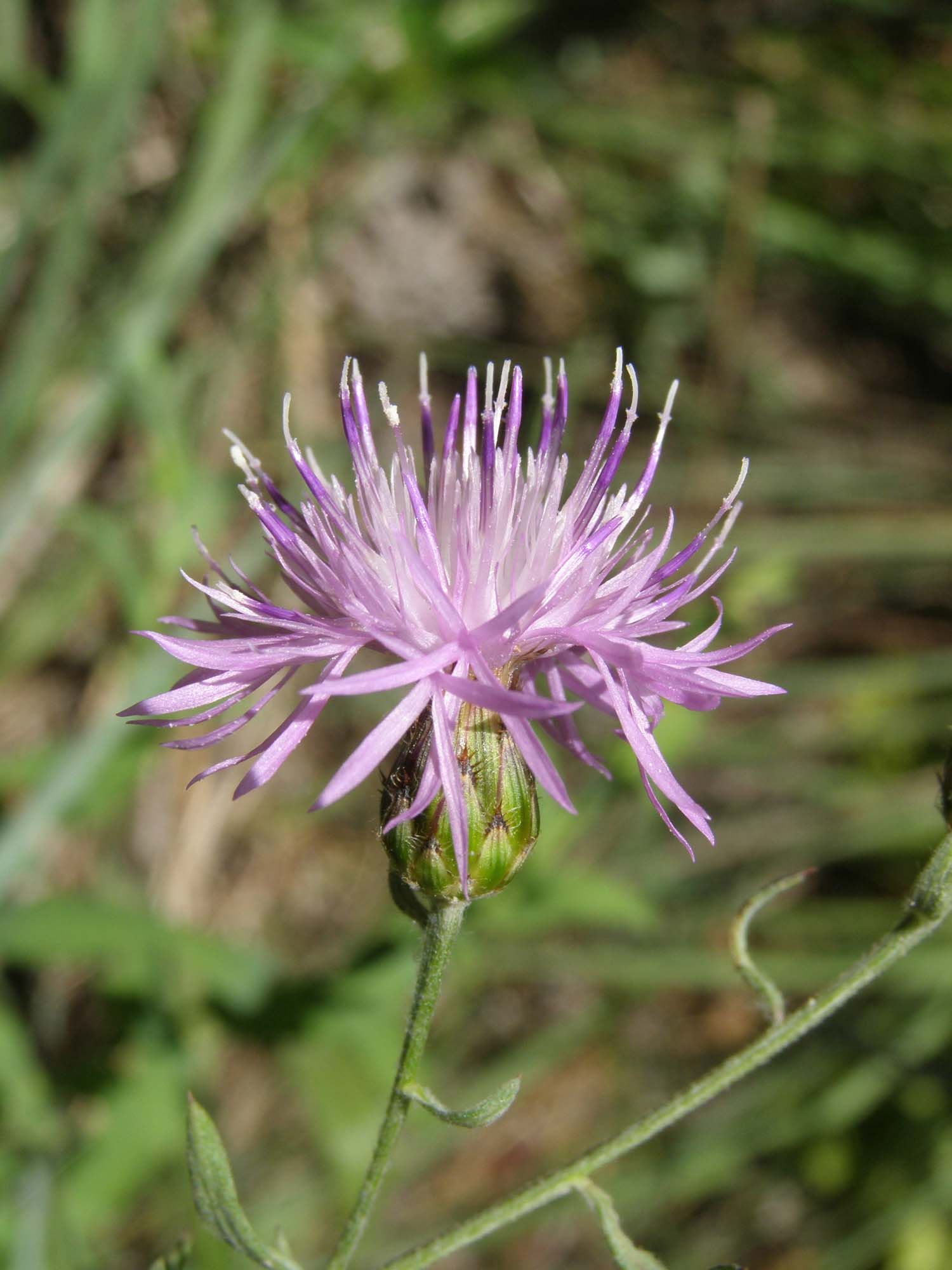} \\

\includegraphics[width=0.14\linewidth, height=1cm]{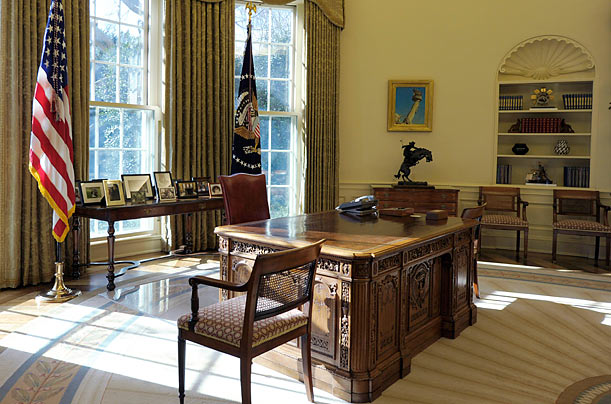} &
\includegraphics[width=0.14\linewidth, height=1cm]{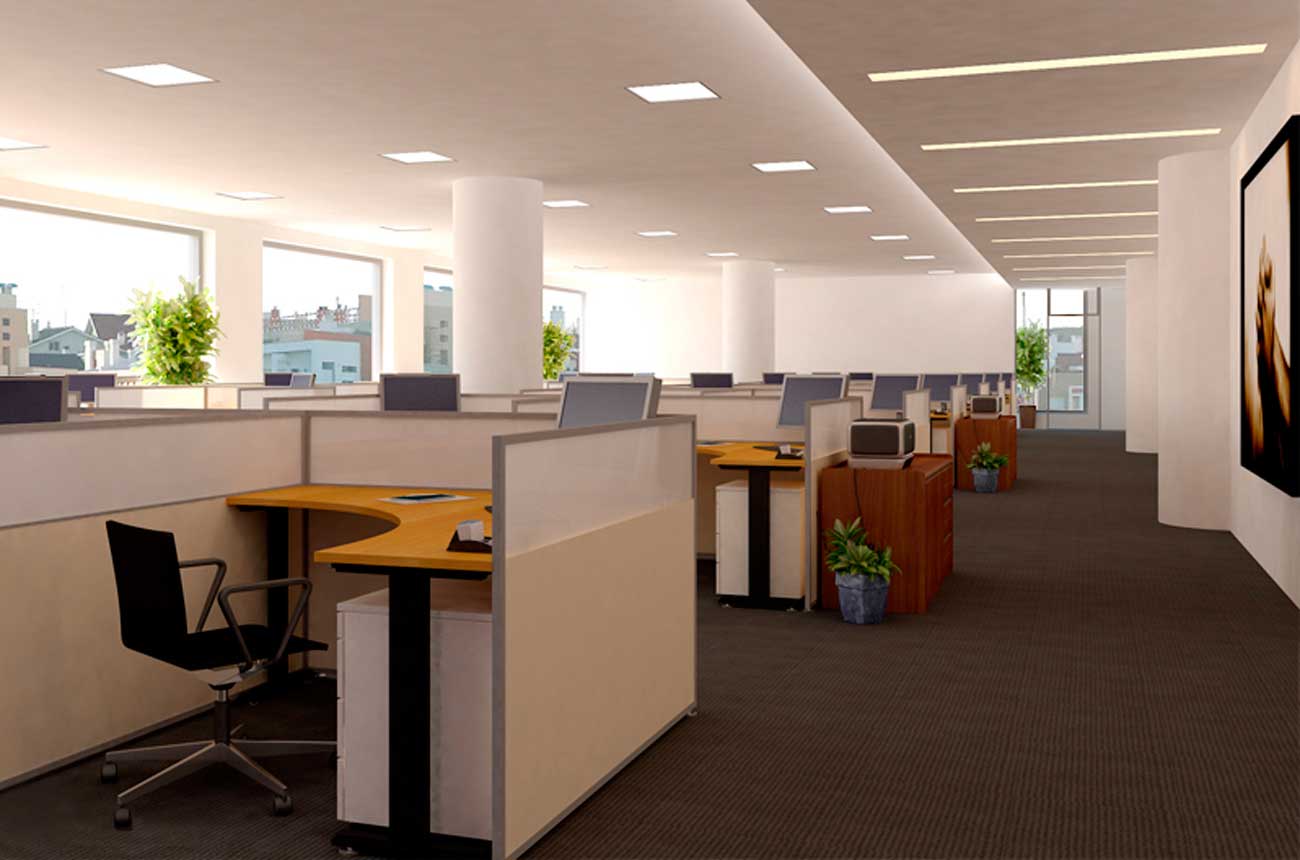} &
\includegraphics[width=0.14\linewidth, height=1cm]{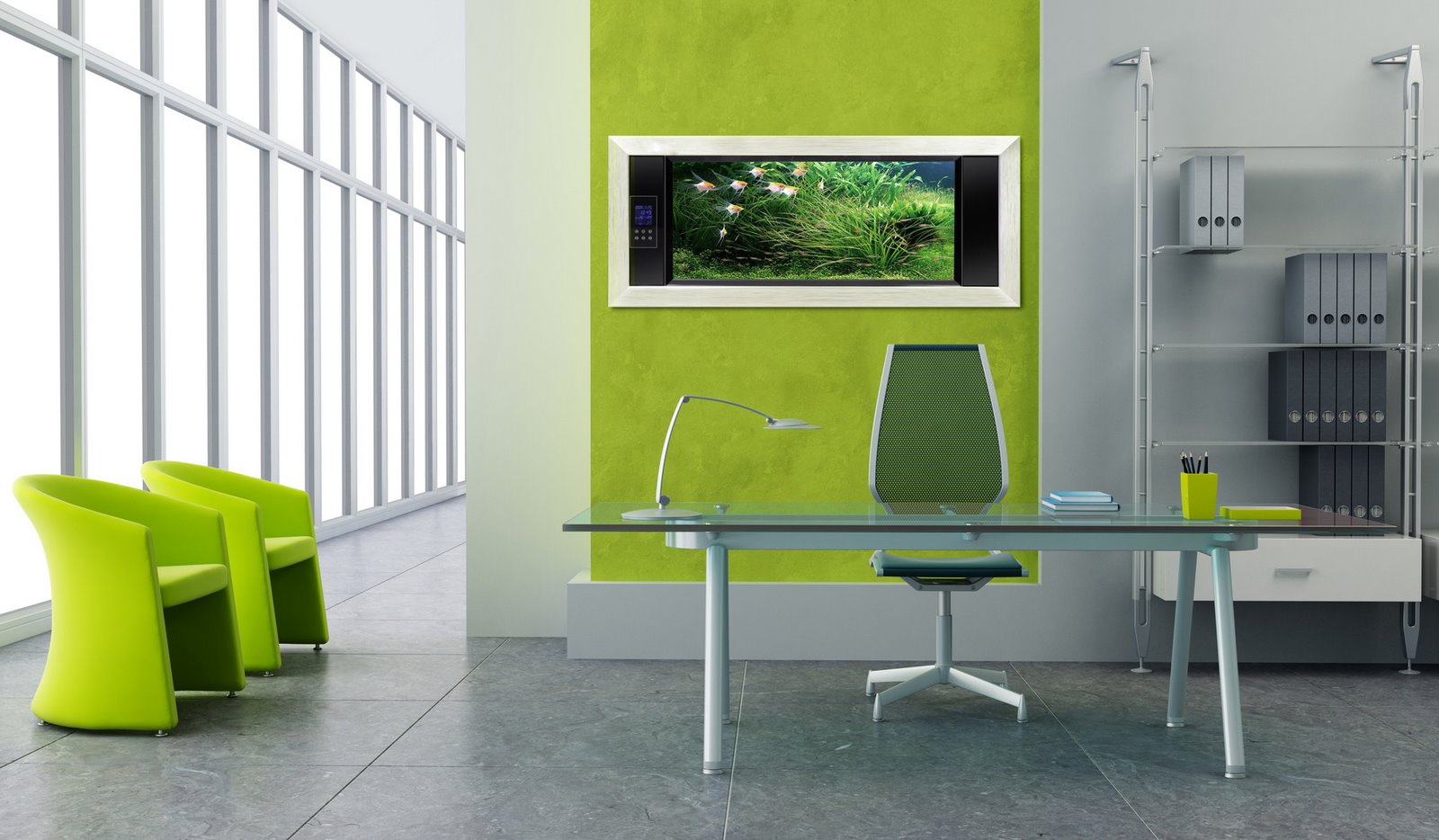} &
\includegraphics[width=0.14\linewidth, height=1cm]{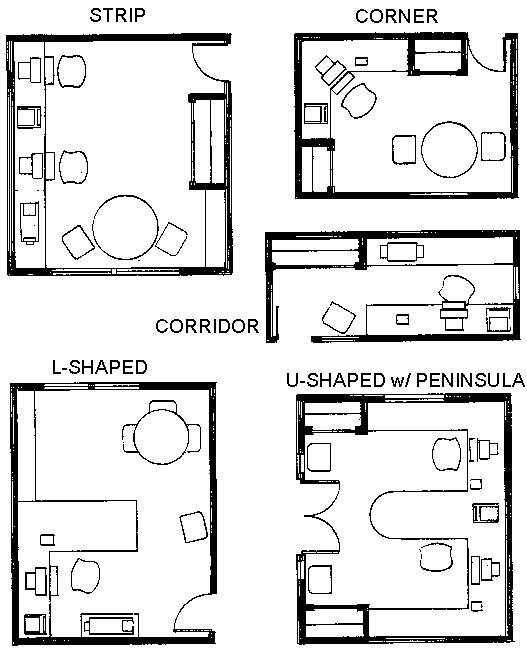} &
\includegraphics[width=0.14\linewidth, height=1cm]{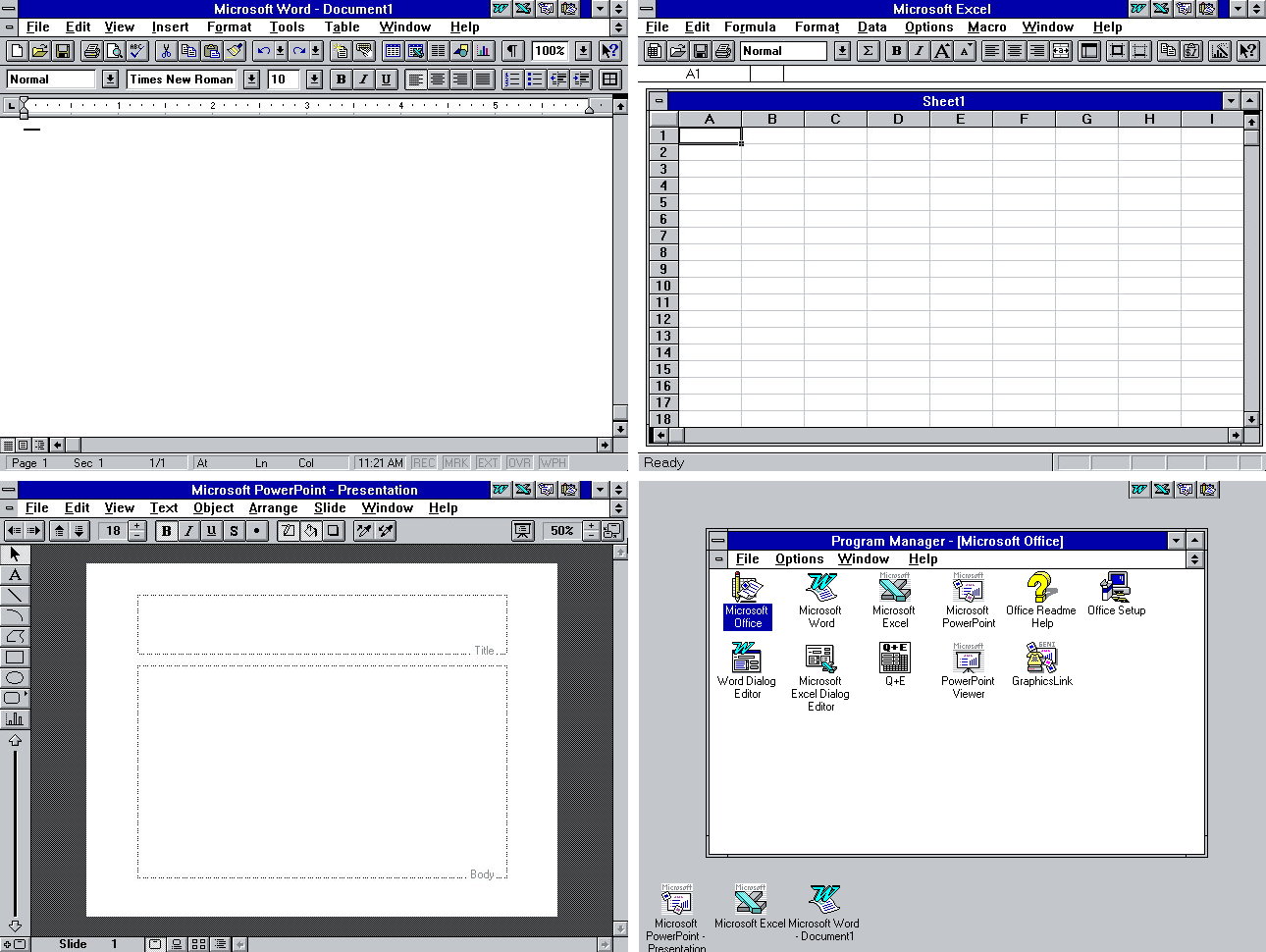} \\
\end{tabular}
\end{center}
\vspace{-0.3cm}
 \caption{Example web images collected for query keywords spotted (top) and office(bottom). Even in the relevant images, the concepts
are observed in different forms requiring
grouping and irrelevant ones to be eliminated.}
\label{fig:examples}
\vspace{-0.2cm}
\end{figure}

Alternatively, images tagged with attribute names are available on the web in large amounts. However, data collected from web inherits all type of challenges due to illumination, reflection, scale, and pose variations as well as camera and compression effects\cite{van2009learning}.  Most importantly, the collection is very noisy with several irrelevant images as well as variety of images corresponding to different characteristic properties of the attribute (Figure\ref{fig:examples}).
Localisation of attributes inside the images arises as another important issue. The region corresponding to the attribute may cover only a fraction of the image, or the same attribute may be in different forms in different parts of an image. 

\begin{figure}[t]
\begin{center}
\fbox{\includegraphics[width=5cm,height=4cm, scale=0.4]{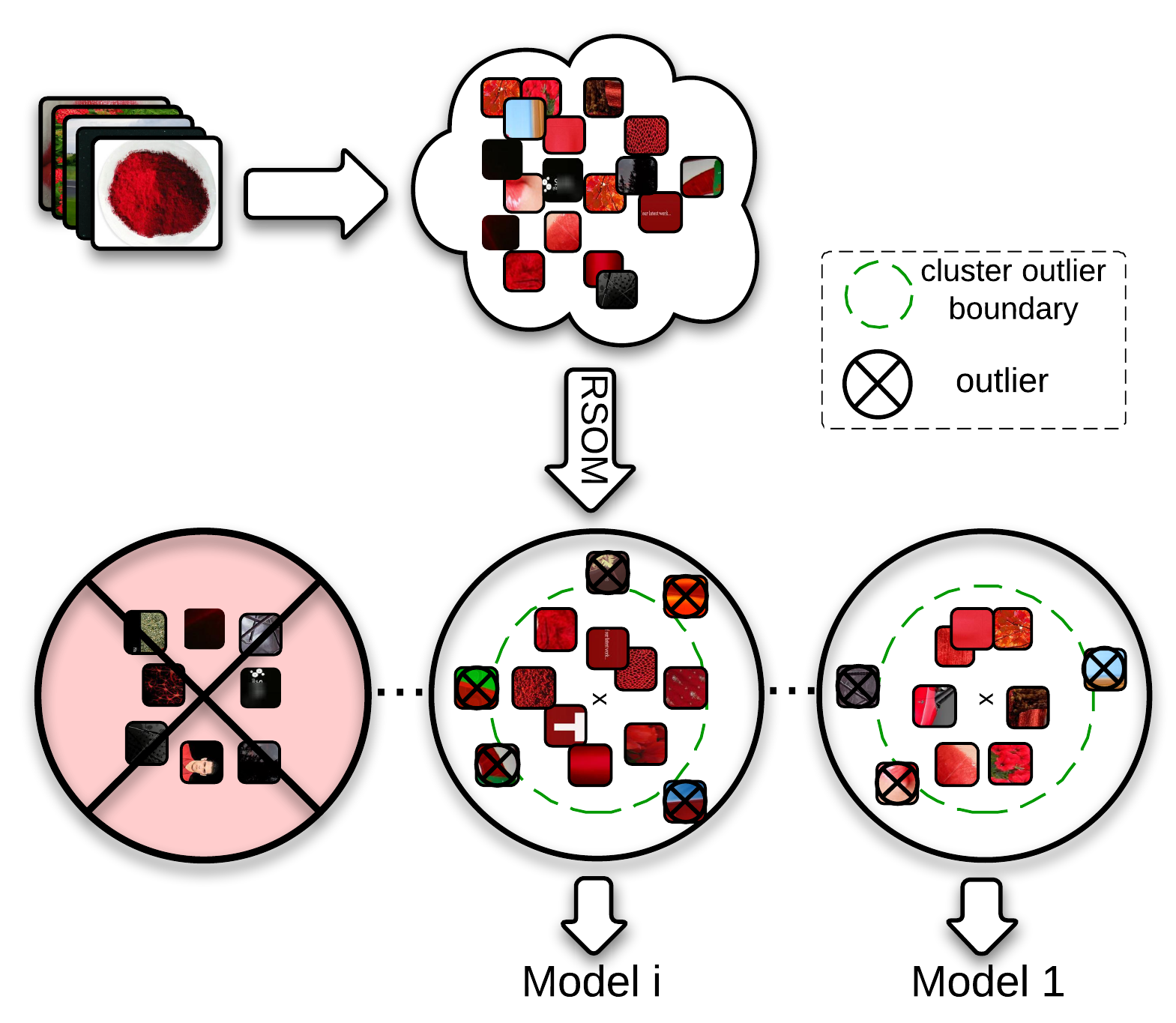}}
\end{center}
   \caption{Overview of the RSOM framework for concept learning.}
\label{fig:overview}
\vspace{-0.2cm}
\end{figure}

For the data collected through querying to be beneficial for automatic learning of attributes, we propose a novel method to obtain an organised collection with irrelevant images removed.  
Our intuition is that, given an attribute category defined by the query word, although the list of images returned is likely to include irrelevant ones, there will be some common characteristics shared among subset of images. Our main idea is to obtain visually coherent subsets, that are possibly corresponding to semantic sub-categories, through clustering and to build models for each sub-category (see Figure\ref{fig:overview}). The model for each attribute category is then a collection of multiple models, each representing a different property of the attribute. 

We aim to answer not only "which attribute is in the image?", but also "where the attribute is?". For this purpose, we consider image patches as the basic units for providing localisation. To retain only the relevant patches that describe the attribute category correctly, during clustering we need to remove outliers, i.e. irrelevant ones. The outliers may resemble to each other while not being similar to the correct category patches resulting in a separate {\bf outlier cluster}.  Alternatively, some outlier patches could be mixed with correct category patches inside {\bf salient clusters} corresponding to relevant ones. These patches, that we refer to as {\bf outlier elements}, should also be removed for the data to be sufficiently clean for learning.

We propose a novel method {\bf Rectifying Self Organizing Maps (RSOM)} which improves the well-known Self Organizing Maps (SOM) \cite{kohonen1997self} through detection and elimination of outliers. The purpose of RSOM is to "rectify" the data by purifying it not only from outlier clusters but also from outlier elements in salient clusters. It is a generic method for capturing category specific characteristics through organising the set of given instances into sub-categories pruned from irrelevant instances. 

Going beyond low-level attributes, RSOM is capable of learning higher level concept, as we show through learning scene concepts. In this case, we treat each image as a single instance, and aim to find groups of images representing a different property of scene category, at the same time by eliminating the ones that are either irrelevant, or poor to sufficiently describe any characteristics.

\section{Related work}
The use of attributes has been the focus of many recent studies~\cite{Farhadi10d.:attribute-centric,Kumar09attributeand,choiadding}.  
In \cite{farhadi2009describing}, Farhadi \etal learn complex attributes (shape, materials, parts) in a fully supervised way focusing on recognition of new types of objects. 
In \cite{lampert2009learning}, for human labelled animal categories, semantic attribute annotations available from studies in cognitive science were used  in a binary fashion for zero-shot learning. In this study, we focus on attribute learning independent of object categories.
Torresani \etal ~\cite{torresani2010efficient} introduce classemes, attributes that do not have specific semantic meanings, but meanings expected to emerge from intersections of properties, and they obtain training data directly from web image search.  
Rastegari \etal \cite{rastegari2012attribute} propose discovering implicit attributes that are not necessarily semantic but preserve category-specific traits through learning discriminative hyperplanes with max-margin and locality sensitive hashing criteria. We learn different intrinsic properties of each attribute through discriminative models obtained from separate clusters that are ultimately combined in a single semantic.
Learning semantic appearance attributes, such as colour, texture and shape, on large scale ImageNet dataset is attacked in~\cite{russakovsky2012attribute} relying on image level human labels using Amazon's Mechanical Turk for supervised learning. We learn attributes from real world images collected from web with no human effort for labelling. Another study on learning colour names from web images is proposed in \cite{van2009learning} where a PLSA based model is used for representing the colour names of pixels. Similar to ours, the approach of Ferrari and Zisserman \cite{ferrari2008learning} considers attributes as patterns sharing some characteristic properties where  basic units are the image segments with uniform appearance. We prefer to work on patch level alternative to pixel level which is not suitable for region level attributes such as texture,  image level which is very noisy, or segment level which is difficult to obtain clearly. 

\section{Rectifying Self Organizing Maps}

{\bf Revisiting Self Organizing Maps (SOM):} Intrinsic dynamics of SOM are inspired from developed animal brain where each part is known to be receptive to different sensory inputs and which has a topographically organized structure\cite{kohonen1997self}. This phenomena, which is called as "receptive field" in visual neural systems \cite{hubel1962receptive}, is simulated with SOM, where neurons are represented by weights that are calibrated to make neurons sensitive to different type of inputs. Elicitation of this structure is furnished by competitive learning approach.

 input with $M$ instances $X=\{x_1,x_2...,x_M\}$. Let  $N=\{n_1,n_2,...,n_K\}$ be the locations of neuron units on the SOM map  and $W=\{w_1,w_2,...,w_K\}$ be the associated weights. 
The neuron whose weight vector is most similar to the input instance  $x_i$ is called as the winner and denoted by $\hat{v}$. The weights of the winner and units in the neighbourhood are adjusted towards the input vector at each iteration $t$ with delta learning rule (Eq.\ref{eq:1}). 
\begin{equation}
\label{eq:1}
w_j^t=w_j^{t-1} + h(n_i, n_{\hat{v}}:\epsilon^t,\sigma^t)[x_i-w_j^{t-1}]\\		
\end{equation}
Update step is scaled by the window function $h(n_i, n_{\hat{v}}:\epsilon^t,\sigma^t)$ for each SOM unit, inversely proportional to the distance to the winner (Eq.\ref{eq:2}). Learning rate $\epsilon$ is a gradually decreasing value, resulting in larger updates at the beginning and finer updates as the algorithm evolves.  $\sigma^t$ defines the neighbouring effect so with the decreasing $\sigma$, neighbour update steps are getting smaller in each epoch. Note that, there are different alternatives for update and windows functions in SOM literature.
\begin{equation}
\label{eq:2}
h(n_i, n_{\hat{v}}:\epsilon^t,\sigma^t) = \epsilon^{t} \exp{\frac{-||n_j - n_{\hat{v}}||^2}{2\sigma^{t^2}}}
\end{equation}

{\bf Clustering and outlier detection with RSOM:}
We introduce excitation scores $E=\{e_1,e_2,\ldots,e_K\}$ where $e_j$, the score for neuron unit $j$, is updated as in Eq.\ref{eq:3}. 
\begin{equation}
\label{eq:3} 
		e_j^t = e_j^{t-1}+\rho^{t}(\beta_j+z_j)	
\end{equation}
As in the original SOM, window function is getting smaller with each iteration. 

Here,  $z_j$ is the activation or win count for the unit $j$, for one epoch. 
$\rho$ is learning solidity scalar that  represents the decisiveness of learning with dynamically increasing value, assuming that later stages of the algorithm has more impact on the definition of salient SOM units. $\rho$ is equal to the inverse of the learning rate $\epsilon$.
$\beta_j$ is the total measure of the activation of $j$th unit in an epoch, caused by all the winners of the epoch but the neuron itself (Eq.\ref{eq:4}). 
\begin{equation}
\label{eq:4}
\beta_{j} = \sum_{\hat{v}=1}^u h(n_j, n_{\hat{v}})z_{\hat{v}}
\end{equation}

At the end of the iterations, normalized $e_j$ is a quality value of a unit $j$. Higher value of $e_j$ indicates that total amount of excitation of the unit $j$ in whole learning period is high thus it is responsive to the given class of instances and it captures notable amount of data. Low excitation values indicate the contrary. RSOM is capable of detecting outlier units via a threshold $\theta$ in the range $[0,1]$.  

Let $C = \{c_1, c_2,\ldots,c_K\}$ be the cluster centres 
corresponding to each unit. $c_j$ is considered to be a {\bf salient cluster} if $e_j \ge \theta$, and an {\bf outlier cluster} otherwise.

The excitation scores $E$ are the measure for saliency of neuron units in RSOM. Given the data belonging to a category, we expect that data is composed of sub-categories that share common properties. For instance {\tt red} images might include darker or lighter tones to be captured by clusters but they are supposed to share a common characteristics of being red. In that sense, for the calculation of the excitation scores we use individual activations of the units as well as the activations as being in a neighbourhood of another unit. Individual activations measure the saliency of being a salient cluster corresponding to a particular sub-category, such as {\tt lighter red}. Neighbourhood activations count the saliency in terms of the shared regularity between sub-categories. If we don't count the neighbourhood effect, some unrelated clusters would be called salient since large number of outlier instances could be grouped in a unit, e.g. noisy white background patches in {\tt red} images.

Outlier instances of salient clusters, namely the {\bf outlier elements} should also be detected. After the detection of outlier neurons, statistics of the distances between neuron weight $w_i$ and its corresponding instance vectors (assuming weights prototyping instances grouped by the neuron) is used as a measure of instance divergence. If the distance between the instance vector $x_j$ and its winner's weight $\hat{w}_i$ is more than the distances of other instances having the same winner, $x_j$ is raised as an outlier element. 
We exploit box plot statistics, similar to \cite{munoz1998self}. If the distance of the instance to its cluster's weight is more than the upper-quartile value, then it is detected as an outlier. The portion of the data, covered by the upper whisker is decided by $\tau$. 

RSOM provides good basis of cleansing of poor instances whereas computing cost is relatively smaller since RSOM is capable of discarding items with one shot of learning phase. Therefore, an additional data cleansing iteration after clustering phase is not required. All the  necessary information (excitation scores, box plot statistics) for outliers is calculated at runtime of learning. Hence, RSOM is suitable for large scale problems.

RSOM is also able to estimate number of intrinsic clusters of the data. We use PCA for that purpose, with defined variation value $\nu$ to be captured by the principle components. Given data and $\nu$, principle components are found and number of principle components describing the data with variation $\nu$ is used as the number of clusters for the further processing of RSOM. If we increase $\nu$, RSOM latches more clusters therefore $\nu$ should be carefully chosen.

 {\bf Discussion of other methods on outlier detection with SOM:}
 \cite{Marsland99amodel,Marsland00} utilise the habitation of the instances. Frequently observed similar instances excites the network to learn some regularities and divergent instances are observed as outliers. 
\cite{harris1993kohonen} benefits from weights prototyping the instances in a cluster. Thresholded distance of instances to the weight vectors are considered as indicator of being outlier. 
In ~\cite{Ypma97noveltydetection}, aim is to have different mapping of activated neuron for the outlier instances.
The algorithm learns the formation of activated neurons on the network for outlier and inlier items with no threshold. It suffers from the generality, with its basic assumption of learning from network mapping.
LTD-KN ~\cite{theofilou2003novelty} performs Kohonen learning rule inversely. An instance activates only the winning neuron as in the usual SOM, but LTD-KN updates winning neuron and its learning windows decreasingly.  

These algorithms only eliminate outlier instances ignoring outlier clusters. RSOM finds outlier clusters as well as the outlier instances in the salient clusters. Another difference of RSOM is the computation cost. Most of outlier detection algorithms model the data and iterate over the data again to label outliers. It is not suitable for large scale data. RSOM has the ability to detect outlier clusters and the items all in the learning phase. Thus, there is no need for learning a model of the data first, then detecting outliers, it is all done in a single pass in our method. With the support of GPGPU programming RSOM scales to large amount of data.
\vspace{-0.1 cm}
\section{Concept learning with RSOM}
\subsection{Learning low-level attributes}

{\bf Data collection and clustering:} We collect web images through querying for colour and texture names. The data is weakly labelled, with the labels given for the entire image, rather than the specific regions. Most importantly, it includes irrelevant images, as well as images with a tiny portion corresponding to the query keyword. 
Each image is densely divided into non-overlapping fixed-size patches to sufficiently capture the required information. We assume that the large volume of the data itself is sufficient to provide instances at various scales and illuminations, and therefore we did not perform any scaling or normalisation. 
The collection of all patches extracted from all images for a single attribute is then given to RSOM to obtain clusters which are likely to capture different characteristics of the attribute

{\bf Training attribute models:} Each cluster obtained through RSOM is used to train a separate classifier for the attribute Positive examples are selected as the members of the cluster and negative instances are selected among the outliers removed by RSOM for that attribute and also among random elements from other attribute categories. We use linear SVM classifier with L1 norm.  

Learned models are used for two different purposes: (i) detection of the attributes on novel images, and (ii) recognition of scenes with the help of learned attribute classifiers.

{\bf Attribute recognition on novel images:} The goal of this task is to label a given image with a single attribute name. 
For this purpose, first we divide the image into grids in three levels using spatial pyramiding \cite{lazebnik2006beyond}. Non-overlapping patches are extracted from each grid in all three levels. Recall that, we have separate classifiers for each salient cluster. We run all of the classifiers on each grid for all patches in all levels. Each grid at each level is labelled by the maximum response classifier among all the outputs for the patches. All of those confidence values are then merged with a weighted sum to a label for the entire image.
\begin{equation}
		D^i = \sum_{l=1}^3 \sum_{n=1}^{N_l} \frac{1}{{2^{3-l}}}h_i e^{-(\hat{x}-x)/2\sigma^2}
		\label{eq:5} 
	\end{equation}
Here,  $N_l$ is the grid number for level $l$ and $h_i$ is the confidence value for grid $i$. We include a Gaussian filter, where $\hat{x}$ is center of the image and $x$ is location of the spatial pyramid grid, to give more priority to the detections around the center of the image for reducing noisy background effect.

{\bf Attribute based scene recognition:}
We use the learned low-level attributes to describe an image for the task of scene recognition. Similar to the first task, we get the confidence values for each grid in three levels of the spatial pyramid. However, rather than using a single value for the maximum classifier output, we use the entire vector for each grid. That is, we keep the confidence values for all the classifiers for each grid. Then, we concatenate these vectors for all grids in all levels to get a single feature vector of size $3xNxK$ for the image, which is then used for scene classification. Here $N$ is the number of grids at each level, and $K$ is the number of different concepts. This rich and high dimensional representation poses good classification performance with simple linear models.

\subsection{Learning higher level concepts}
To show that RSOM is capable of being generalised to higher level concepts, we collected images for scene categories from web to learn these concepts. 
In this case, we use the entire images as instances, and aim to discover group of images each representing a different property of the scene category. These clusters are then used as models similar to the attribute learning.  Specifically, we perform experiments for scene classification for 15 scene categories as in \cite{lazebnik2006beyond}. Note that, we do not use any manually labelled training set, but directly the noisy web images which are pruned and organised by RSOM. This task is also different than the use of low-level attributes for scene recognition, in this case we learn the scene concept directly without requiring any other information.

\section{Experiments}
\subsection{Qualitative evaluation of clusters}
\begin{figure}
\begin{center}
\fbox{\includegraphics[width=8cm,height=6cm, scale=0.4]{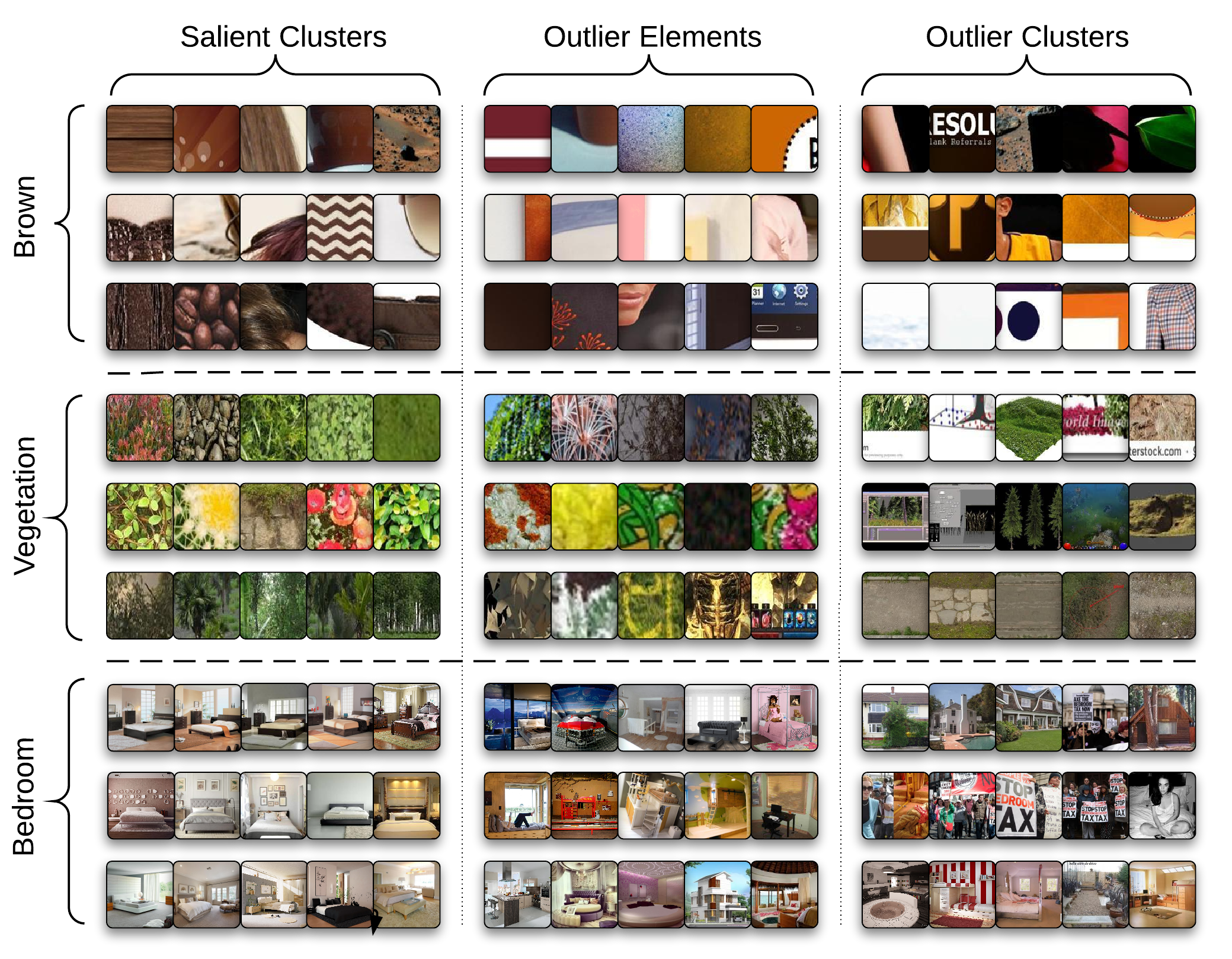}}
\end{center}
   \caption{For two low level concepts \textit{brown} and \textit{vegetation} and one high level concept \textit{bedroom},  random images detected as (i) elements of salient clusters, (ii) elements of outlier clusters, and (iii) outlier elements in salient clusters. }
\label{fig:short}
\end{figure}

As Figure\ref{fig:short} depicts, RSOM captures different characteristics of concepts in separate salient clusters, while eliminating outlier clusters that group irrelevant images which are coherent among themselves, as well as outlier elements wrongly mixed with the elements of salient clusters. 

\subsection{Implementation details}
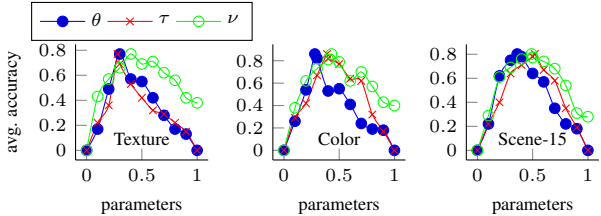
\begin{figure}
\centering
\mbox{

		\pgfplotsset{every axis legend/.append style={at={(0.5,1.03)},anchor=south},scale=0.45,width=5.5cm,height=5cm,compat=1.3,every axis/.append style={
extra description/.code={font=\scriptsize,
\node at (0.5,0.18) {Texture};
}}}
		\begin{tikzpicture}
		\begin{axis}[
			enlargelimits=0.1,
			xlabel={parameters},
			ylabel={avg. accuracy},
			ytick={0,0.2,0.4,0.6,0.8,1},
			xmin=0,
			x tick label style={font= \scriptsize},
			x label style={font= \scriptsize},
			y tick label style={font=\scriptsize},
			y label style={font=\scriptsize},
			legend style={font=\scriptsize},
			axis x line*=none,
        		axis y line*=none,
        		legend columns=4
		]
		
		\addplot+[smooth,mark=*,color=blue]
		coordinates
		{(0,0) (0.1,0.17) (0.2,0.49) (0.3,0.77)  (0.4,0.57) (0.5,0.55) (0.6,0.42) (0.7,0.28) (0.8,0.17) (0.9,0.13) (1,0)};
		\addlegendentry{$\theta$}
		
		\addplot+[smooth,mark=x, color=red]
		coordinates
		{(0,0) (0.1,0.22) (0.2,0.36) (0.28, 0.77) (0.3,0.69) (0.4,0.53) (0.5,0.42) (0.6,0.32) (0.7,0.29) (0.8,0.22) (0.9,0.15) (1,0)};
		\addlegendentry{$\tau$}
		
		\addplot+[smooth,mark=o, color=green]
		coordinates
		{(0,0) (0.1,0.43) (0.2,0.57) (0.3,0.66) (0.4,0.77) (0.5,0.69) (0.6,0.71) (0.7,0.62) (0.8,0.56) (0.9,0.42) (1,0.38)};
		\addlegendentry{$\nu$}
		
		\end{axis}
		\end{tikzpicture}
	
	\hspace{-0.6cm}

		\pgfplotsset{width=5.5cm,height=5cm,compat=1.3,every axis/.append style={
extra description/.code={font=\scriptsize,
\node at (0.5,0.18) {Color};
}}}
		\begin{tikzpicture}
		\begin{axis}[
			enlargelimits=0.1,
			xlabel={parameters},
			ytick={0,0.2,0.4,0.6,0.8,1},
			xmin=0,
			x tick label style={font= \scriptsize},
			x label style={font= \scriptsize},
			y tick label style={font=\scriptsize},
			y label style={font=\scriptsize},
			legend style={font=\scriptsize},
			axis x line*=none,
        		axis y line*=none,
		]
		
		\addplot+[smooth,mark=*,color=blue]
		coordinates
		{(0,0) (0.1,0.26) (0.2,0.54) (0.28,0.86) (0.3,0.82) (0.4,0.53) (0.5,0.55) (0.6,0.41) (0.7,0.24) (0.8,0.19) (0.9,0.18) (1,0)};
	
		\addplot+[smooth,mark=x, color=red]
		coordinates
		{(0,0) (0.1,0.29) (0.2,0.42) (0.3,0.67) (0.39,0.86) (0.4,0.82) (0.5,0.77) (0.6,0.64) (0.7,0.62) (0.8,0.32) (0.9,0.17) (1,0)};
		
		\addplot+[smooth,mark=o, color=green]
		coordinates
		{(0,0) (0.1,0.38) (0.2,0.62) (0.3,0.73) (0.4,0.81) (0.43,0.86) (0.5,0.79) (0.6,0.62) (0.7,0.70) (0.8,0.57) (0.9,0.43) (1,0.40)};

		r
		\end{axis}
		\end{tikzpicture}

\hspace{-0.1cm}

		\pgfplotsset{width=5.5cm,height=5cm,compat=1.3,every axis/.append style={
extra description/.code={font=\scriptsize,
\node at (0.5,0.18) {Scene-15};
}}}
		\begin{tikzpicture}
		\begin{axis}[
			enlargelimits=0.1,
			xlabel={parameters},
			ytick={0,0.2,0.4,0.6,0.8,1},
			xmin=0,
			x tick label style={font= \scriptsize},
			x label style={font= \scriptsize},
			y tick label style={font=\scriptsize},
			y label style={font=\scriptsize},
			legend style={font=\scriptsize},
			axis x line*=none,
        		axis y line*=none,
		]
		
		\addplot+[smooth,mark=*,color=blue]
		coordinates
		{(0,0) (0.1,0.22) (0.2,0.62)  (0.3,0.75) (0.36,0.802)  (0.4,0.78) (0.5,0.64) (0.6,0.57) (0.7,0.35) (0.8,0.22) (0.9,0.18) (1,0)};
		
		\addplot+[smooth,mark=x, color=red]
		coordinates
		{(0,0) (0.1,0.22) (0.2,0.40) (0.3,0.64)  (0.4,0.71) (0.5,0.78)(0.52,0.802)  (0.6,0.67) (0.7,0.58) (0.8,0.35) (0.9,0.19) (1,0)};
		
		\addplot+[smooth,mark=o, color=green]
		coordinates
		{(0,0) (0.1,0.29) (0.2,0.61) (0.3,0.69) (0.4,0.76) (0.49,0.802) (0.5,0.773) (0.6,0.74) (0.7,0.68) (0.8,0.54) (0.9,0.31) (1,0.28)};
		
		\end{axis}
		\end{tikzpicture}
}
 \caption{Effect of parameters on average accuracy. For each parameter, the other two are fixed at their optimum values.}
	\label{fig:theta vs accuracy}
\end{figure}

Parameters of RSOM are selected on a small held-out set. Figure\ref{fig:theta vs accuracy} depicts the effect of parameters $\theta$, $\tau$ and $\nu$. For each parameter the other two are fixed at the optimum value obtained through cross-validation.\\
SVM parameters are also selected with 10-fold cross validation and grid-search. We end the search process when the current accuracy is less than the average accuracy of the 5 to 10 step back.  
Our RSOM implementation is powered by GPGPU programming over CUDA environment, resulting in a large time reduction. 

\subsection{Attribute learning} 
{\bf Datasets: } We collected images from Google for 11 distinct colours as in \cite{van2009learning} and 13 textures. We included the terms "colour" and "texture" in the queries, such as ''red colour", or "wooden texture",  to reduce the chance of semantic mismatching. For each attribute, 500 images are collected and patches are extracted from each image.  
To test the results on a human labelled dataset, we use Ebay dataset  provided by \cite{van2009learning} which has labels for the pixels in cropped regions. Unlike \cite{van2009learning}, we didn't apply gamma correction.

{\bf Tasks :} Evaluation of our framework is performed over two different tasks: (i) detection of the learned attributes on novel images and (ii) recognition of scenes using learned attributes. To evaluate the first task, we use three different datasets. The first dataset is Bing Search Images curated by ourselves from the top 35 images returned with some pruning to eliminate some semantic mismatches.  Second dataset is Google Colour Images \cite{van2009learning} previously used by \cite{van2009learning}  for learning colour attributes. We used the data only for testing. The last dataset is sample annotated images from ImageNet \cite{russakovsky2012attribute} for 25 attributes. %
The second task on scene recognition is performed on MIT-indoor \cite{quattoni2009recognizing}, and Scene-15 \cite{lazebnik2006beyond} datasets.

{\bf Representation:}
For color concepts we use 10x20x20 bins Lab colour histograms and for texture concepts we also include BoW representation for densely sampled SIFT [REF] features with 4000 words. 

We keep the feature dimensions high to utilise from the over-complete representations of the instances when we apply L1 norm linear SVM classifier. %
We divide the training images into 100x100 non-overlapping patches. We did not perform any scaling or normalization and capture different varieties with crowd of the patches. For testing, we divide the images into 21 grids at each of the three levels.

{\bf Results:} Figure \ref{fig:accuracyPlot} compares the accuracy of the proposed method ({\bf RSOM}) with three other methods on the task of attribute learning. As a baseline method ({\bf BL}), we use all the images returned for the concept query to train a single model. As the results show, the performance is very low suggesting that the data should be organised to capture the intra-class variations.  As two other methods for clustering the data, we used k-means ({\bf KM}) and original SOM algorithm ({\bf SOM}), and again train different models for each clusters. The low results support the need for pruning of the data through outlier elimination. Results also show that, on novel test sets with images having different characteristics than the images used in training, RSOM can still perform very well on learning of attributes.

Note that, on ImageNet dataset, we obtained 37.4\% accuracy compared to 36.8\% of Russakovsky and Fei-Fei\cite{russakovsky2012attribute}.

Our method is also utilised for retrieving images on EBAY dataset as in \cite{van2009learning}. We utilise RSOM with patches obtained from the entire images ({\bf RSOM}) as well as from the masks provided by \cite{van2009learning} ({\bf RSOM-M}).  As shown in Table\ref{table:PLSAvsRSOM}, even without masks RSOM is comparable to the performance of the PLSA based method of \cite{van2009learning}, and with the same setting RSOM outperforms the PLSA based method.  	

On the task of scene recognition with learned attributes, we compare our method ({\bf RSOM-A}) with state-of-the-art methods on MIT-indoor \cite{quattoni2009recognizing} and Scene-15  \cite{lazebnik2006beyond} datasets. Our method performs competitively with\cite{liharvesting} while using shorter feature vectors, and outperforms the others. 

\definecolor{rsomColor}{HTML}{D7191C}
\definecolor{somColor}{HTML}{FDAE61}
\definecolor{kmColor}{HTML}{ABDDA4}
\definecolor{blColor}{HTML}{666699}

\pgfplotsset{width=8cm, height=3.5cm,compat=1.3}
\begin{figure}
\begin{center}
\begin{tikzpicture}
\begin{axis}[
			ybar,
			enlargelimits=0.25,
			xtick=data,
			xticklabels={Bing,Google \cite{van2009learning} ,ImageNet \cite{russakovsky2012attribute},EBAY \cite{van2009learning}},
			ylabel={accuracy},
			legend style={font=\scriptsize},
			x tick label style={font= \scriptsize,rotate=90, anchor=north},
			y label style={font=\footnotesize},
			legend style={
					area legend,
					at={(0.5,1.1)},
					anchor=north,
					rotate=-90,
					legend columns=4,
					draw=none},					
			x tick label style={rotate=-90},
			bar width=3.5pt,	
		    area legend,
		    axis x line*=none,
        		axis y line*=none,
]

\addplot [fill=rsomColor] coordinates
{(0,0.82) (1,0.70) (2,0.37) (3,0.81)};

\addplot [fill=somColor] coordinates
{(0,0.56) (1,0.31) (2,0.21) (3,0.68)};

\addplot [fill=kmColor] coordinates
{(0,0.57) (1,0.30) (2,0.22) (3,0.69)};

\addplot [fill=blColor] coordinates
{(0,0.31) (1,0.25) (2,0.17) (3,0.56)};
\legend{RSOM,SOM,KM,BL}
\end{axis}
\end{tikzpicture}
\end{center}
\vspace{-0.3cm}
\caption{Attribute recognition performances on novel images.}
\label{fig:accuracyPlot}
\end{figure}
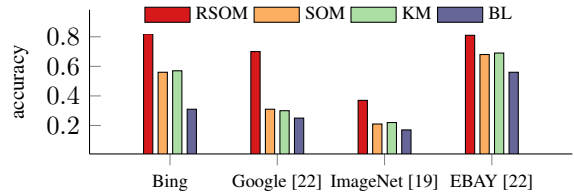

\begin{table}
\footnotesize
\begin{center}
\begin{tabular}{|c|c|c|c|}
\hline 
Method & RSOM-M & RSOM & PLSA-reg \cite{van2009learning}.\\ 
\hline 
cars & 0.97 & 0.92 & 0.93 \\ 
\hline 
shoes & 1.0 & 0.97 & 0.99 \\ 
\hline 
dresses & 1.0 & 1.0 & 0.99 \\ 
\hline 
pottery & 0.98 & 0.92 & 0.94 \\ 
\hline 
\hline
overall & 0.99 & 0.95 & 0.96 \\ 
\hline 
\end{tabular} 
\end{center}
\caption{Equal Error Rates on EBAY dataset for image retrieval using the configuration of \cite{van2009learning}.  RSOM does not utilise the image masks used in \cite{van2009learning}, while RSOM-M does.}
\label{table:PLSAvsRSOM}
\end{table}

\begin{table}
\footnotesize
\begin{center}
\begin{tabular}{|c|c|c|}
\hline 
Method & MIT-indoor \cite{quattoni2009recognizing} & Scene-15  \cite{lazebnik2006beyond} \\ 
\hline 
RSOM-A 								& 46.2\% & 82.7\%   \\ 
\hline
RSOM-S								& - & 80.7\%		\\
\hline
RSOM-S+HM							& - & 81.3\% \\ 
\hline 
Li \etal \cite{liharvesting} VQ 		& 47.6\% & 82.1\%  \\ 
\hline
Pandey \etal \cite{pandey2011scene} 	& 43.1\% & - \\
\hline
Kwitt \etal \cite{kwitt2012scene} 	& 44\% & 82.3\%\\
\hline
Lazebnik \etal \cite{lazebnik2006beyond} & - & 81.0\%\\
\hline
\end{tabular} 
\end{center}
\caption{Comparison of our methods on scene recognition in relation to state-of-the-art studies on MIT-Indoor \cite{quattoni2009recognizing} and Scene-15  \cite{lazebnik2006beyond} datasets. }
\label{liharvestVSrsom}
\end{table}

\subsection{Learning concepts of scene categories}
As an alternative to recognising scenes through the learned low-level attributes, we directly learn higher level concepts for scene categories. We focus on learning 
15 scene concepts used in \cite{lazebnik2006beyond}, through collecting images from web for these concepts.  We have shown that, our method is competitive with the state-of-the-art studies without requiring any supervised training. We made a slight change on our original RSOM implementation for recognising scene concepts (which we refer to as {\bf RSOM-S}) by  finding the hard negatives at the first classification and using them in another classification (we refer to this new method as {\bf RSOM-S-HM}). As the results in Figure\ref{fig:accuracyPlot2} show, we achieve better performances than the state-of-the-art studies with this simple addition, still without requiring any supervisory input.
 
\pgfplotsset{width=8.75cm, height=3.5cm,compat=1.3}
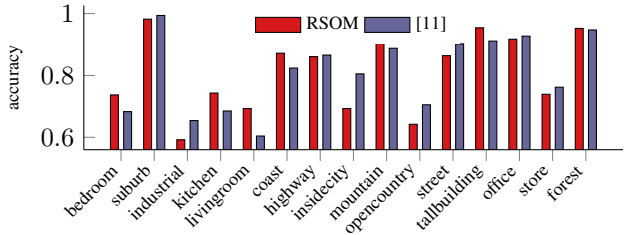
\begin{figure}
\begin{center}
\begin{tikzpicture}
\begin{axis}[
			ybar,
			enlargelimits=0.08,
			xtick=data,
			xticklabels={bedroom,suburb,industrial, kitchen, livingroom, coast, highway, insidecity, mountain, opencountry, street, tallbuilding, office, store, forest},
			ylabel={accuracy},
			legend style={font=\scriptsize},
			x tick label style={font= \scriptsize,rotate=135, anchor=east},
			y label style={font=\scriptsize},
			legend style={
					area legend,
					at={(0.5,1)},
					anchor=north,
					rotate=-90,
					legend columns=2,
					draw=none},					
			x tick label style={rotate=-90},
			bar width=3pt,	
		    area legend,
		    axis x line*=none,
        		axis y line*=none,
]

\addplot [fill=rsomColor] coordinates
{(0,0.737) (1,0.982) (2,0.592) (3,0.743) (4,0.693) (5,0.872) (6,0.861)(7,0.693)(8,0.947)(9,0.642)(10,0.864)(11,0.954)(12,0.917)(13,0.739)(14,0.952)};

\addplot [fill=blColor] coordinates
{(0,0.683) (1,0.994) (2,0.654) (3,0.685) (4,0.604) (5,0.824) (6,0.866)(7,0.805)(8,0.888)(9,0.705)(10,0.902)(11,0.911)(12,0.927)(13,0.762)(14,0.947)};

\legend{RSOM,\cite{lazebnik2006beyond}}

\end{axis}
\end{tikzpicture}
\end{center}
\vspace{-0.3cm}
\caption{Comparisons on Scene-15 dataset. Overall accuracy is 81.3\% for RSOM-S+HM , versus 81\% for \cite{lazebnik2006beyond} . Classes "industrial", "insidecity", "opencountry" results very noisy set of web images, hence trained models are not strong enough as might be observed from the chart.}
\label{fig:accuracyPlot2}
\vspace{-0.1cm}
\end{figure}

\section{Conclusion}
In this work we propose Rectifying Self Organizing Maps that is akin to SOM with clustering properties and novel with respect to outlier detection dynamics. 
We use RSOM for weakly supervised learning of visual concepts from large scale noisy web data. Multiple classifiers are built for each attribute from clusters pruned from outliers, to have each classifier sensitive to a different visual variation. 
Our experiments show that we are able to capture low level concepts on novel images and have a good basis for higher level recognition tasks like scene recognition with inexpensive setting. We also show that we can directly learn higher level concepts.
As the future work, this framework will be extended to capture more localized concepts like objects, and will also be applied to learn concepts from videos.
 
{\small
\bibliographystyle{ieee}
\bibliography{RSOM-Golge-CVPR2014}

\begin{thebibliography}{10}\itemsep=-1pt

\bibitem{choiadding}
J.~Choi, M.~Rastegari, A.~Farhadi, and L.~S. Davis.
\newblock Adding unlabeled samples to categories by learned attributes.
\newblock {\em CVPR}, 2013.

\bibitem{Farhadi10d.:attribute-centric}
A.~Farhadi, I.~Endres, and D.~Hoiem.
\newblock D.: Attribute-centric recognition for crosscategory generalization.
\newblock {\em CVPR}, 2010.

\bibitem{farhadi2009describing}
A.~Farhadi, I.~Endres, D.~Hoiem, and D.~Forsyth.
\newblock Describing objects by their attributes.
\newblock {\em CVPR}, 2009.

\bibitem{ferrari2008learning}
V.~Ferrari and A.~Zisserman.
\newblock Learning visual attributes.
\newblock {\em NIPS}, 2008.

\bibitem{harris1993kohonen}
T.~Harris.
\newblock A kohonen som based, machine health monitoring system which enables
  diagnosis of faults not seen in the training set.
\newblock {\em Neural Networks. IJCNN'93-Nagoya.}, 1993.

\bibitem{hubel1962receptive}
D.~H. Hubel and T.~N. Wiesel.
\newblock Receptive fields, binocular interaction and functional architecture
  in the cat's visual cortex.
\newblock {\em The Journal of physiology}, 1962.

\bibitem{kohonen1997self}
T.~Kohonen.
\newblock {\em Self-organizing maps}.
\newblock Springer, 1997.

\bibitem{Kumar09attributeand}
N.~Kumar, A.~C. Berg, P.~N. Belhumeur, and S.~K. Nayar.
\newblock Attribute and simile classifiers for face verification.
\newblock {\em ICCV}, 2009.

\bibitem{kwitt2012scene}
R.~Kwitt, N.~Vasconcelos, and N.~Rasiwasia.
\newblock Scene recognition on the semantic manifold.
\newblock {\em ECCV}, 2012.

\bibitem{lampert2009learning}
C.~H. Lampert, H.~Nickisch, and S.~Harmeling.
\newblock Learning to detect unseen object classes by between-class attribute
  transfer.
\newblock {\em CVPR}, 2009.

\bibitem{lazebnik2006beyond}
S.~Lazebnik, C.~Schmid, and J.~Ponce.
\newblock Beyond bags of features: Spatial pyramid matching for recognizing
  natural scene categories.
\newblock {\em CVPR}, 2006.

\bibitem{liharvesting}
Q.~Li, J.~Wu, and Z.~Tu.
\newblock Harvesting mid-level visual concepts from large-scale internet
  images.
\newblock {\em CVPR}, 2013.

\bibitem{Marsland99amodel}
S.~Marsland, U.~Nehmzow, and J.~Shapiro.
\newblock A model of habituation applied to mobile robots.
\newblock {\em Proceedings of Towards Intelligent Mobile Robots}, 1999.

\bibitem{Marsland00}
S.~Marsland, U.~Nehmzow, and J.~Shapiro.
\newblock {Novelty Detection for Robot Neotaxis}.
\newblock {\em Proceedings 2nd NC}, 2000.

\bibitem{munoz1998self}
A.~Mu{\~n}oz and J.~Muruz{\'a}bal.
\newblock Self-organizing maps for outlier detection.
\newblock {\em Neurocomputing}, 1998.

\bibitem{pandey2011scene}
M.~Pandey and S.~Lazebnik.
\newblock Scene recognition and weakly supervised object localization with
  deformable part-based models.
\newblock {\em ICCV}, 2011.

\bibitem{quattoni2009recognizing}
A.~Quattoni and A.~Torralba.
\newblock Recognizing indoor scenes.
\newblock {\em CVPR}, 2009.

\bibitem{rastegari2012attribute}
M.~Rastegari, A.~Farhadi, and D.~Forsyth.
\newblock Attribute discovery via predictable discriminative binary codes.
\newblock {\em ECCV}, 2012.

\bibitem{russakovsky2012attribute}
O.~Russakovsky and L.~Fei-Fei.
\newblock Attribute learning in large-scale datasets.
\newblock {\em Trends and Topics in Computer Vision}, 2012.

\bibitem{theofilou2003novelty}
D.~Theofilou, V.~Steuber, and E.~D. Schutter.
\newblock Novelty detection in a kohonen-like network with a long-term
  depression learning rule.
\newblock {\em Neurocomputing}, 2003.

\bibitem{torresani2010efficient}
L.~Torresani, M.~Szummer, and A.~Fitzgibbon.
\newblock Efficient object category recognition using classemes.
\newblock {\em ECCV}, 2010.

\bibitem{van2009learning}
J.~Van De~Weijer, C.~Schmid, J.~Verbeek, and D.~Larlus.
\newblock Learning color names for real-world applications.
\newblock {\em Image Processing, IEEE}, 2009.

\bibitem{Ypma97noveltydetection}
A.~Ypma, E.~Ypma, and R.~P. Duin.
\newblock Novelty detection using self-organizing maps.
\newblock {\em In Proc. of ICONIP'97}, 1997.

\end{thebibliography}
} 
\end{document}